# Sequence to Sequence Networks for Roman-Urdu to Urdu Transliteration

*Mehreen Alam, Sibt ul Hussain*

*mehreen.alam@nu.edu.pk, sibtul.hussain@nu.edu.pk*

*Reveal Lab, Computer Science Department, NUCES*

*Islamabad, Pakistan*

*Abstract*— Neural Machine Translation models have replaced the conventional phrase based statistical translation methods since the former takes a generic, scalable, data-driven approach rather than relying on manual, hand-crafted features. The neural machine translation system is based on one neural network that is composed of two parts, one that is responsible for input language sentence and other part that handles the desired output language sentence. This model based on encoder-decoder architecture also takes as input the distributed representations of the source language which enriches the learnt dependencies and gives a warm start to the network. In this work, we transform Roman-Urdu to Urdu transliteration into sequence to sequence learning problem. To this end, we make the following contributions. We create the first ever parallel corpora of Roman-Urdu to Urdu, create the first ever distributed representation of Roman-Urdu and present the first neural machine translation model that transliterates text from Roman-Urdu to Urdu language. Our model has achieved the state-of-the-art results using BLEU as the evaluation metric. Precisely, our model is able to correctly predict sentences up to length 10 while achieving BLEU score of 48.6 on the test set. We are hopeful that our model and our results shall serve as the baseline for further work in the domain of neural machine translation for Roman-Urdu to Urdu using distributed representation.

## I. INTRODUCTION

Deep neural networks have shown a remarkable performance in many applications from computer vision [1] to speech recognition [2]. Specifically, in the field of natural language processing, the conventional phrase based statistical machine translation systems have been superseded by the neural machine translation methods. To the best of our knowledge, no significant work has been done to apply deep learning techniques for any of the variety of natural language processing tasks in Urdu language[1]. To date, conventional natural language processing methods are being employed for Urdu language which limit the extent to which the performance can be achieved. Their reliance on hand-crafted features and manual annotations restricts not only their scalability for larger datasets but also their capacity to handle varying complexities of the language. In this paper, neural machine translation based on deep learning has been applied to get transliteration from Roman-Urdu to Urdu with distributed representations of both the languages.

Neural Machine Translation is based on the concept of encoder-decoder where encoder and decoder are both based on the sequence of multiple Recurrent Neural Network (RNN) cells [3]. The encoder maps a variable-length source sequence to a fixed-length vector, and the decoder maps the vector representation back to a variable-length target sequence. The two networks are trained jointly to maximize the conditional probability of the target sequence given a source sequence [3]. The longer the sequence, better the model is able to capture long-term dependencies in the language. The source language is fed to the encoder and target language to the decoders. The size of encoder-decoder and the length of input and output sequences need not be same. This gives the flexibility to accommodate the heterogeneity inherent in any two languages. It is also possible that the input and the output languages may differ in various ways, for example: being right-to-left or left-to-right language; on degree of morphological richness; or on vocabulary size or alignment.

For our work, we have chosen LSTM based cells for our encoder and decoders since they are known to overcome the problem of vanishing gradients [4] and can also capture longer dependencies in the sentence. Precisely, to let our model learn the richer contextual and sequential structure we have used multi-layer (three layers) LSTM based encoder and decoder. As a warm start and to help the model better learn the relationships between the two languages, we feed the model with the vector representation of each word where each vector has been generated by using word2vec [5] for both the languages. We trained our model on Roman-Urdu to Urdu parallel corpus of size 0.1 million lines containing 1.74 million words for each corpus. This includes 20K unique Roman-Urdu words and 19K unique Urdu words.

---

[1] Google's GBoard has released a Roman-Urdu to Urdu transliteration application but the technique used is yet unknown.

Quantitative analysis of the accuracy of the transliteration was done using Bilingual Evaluation Understudy (BLEU), proposed by [6]. Our model was able to achieve a BLEU score 50.68 and 48.6 on train and test data respectively. For further details, please refer to the experiments section.

Rest of the paper is organized as follows: section II surveys the related work. Detailed explanation of the model, experimental settings, data collection, pre-processing and comprehensive analysis with discussion on results are presented in Section III. Section IV lists the conclusions while future work is discussed Section V.

## II. MOTIVATION AND RELATED WORK

Since its inception, sequence to sequence (seq2seq) network has become the most successful technique based on neural networks for mapping an input sequence to predict its corresponding sequence. It has been applied to solve a variety of problems: conversational modeling [7], handwriting generation, question answering, speech recognition [42], music, protein secondary structure prediction, text-to-speech, polyphonic music modeling, speech signal modeling and machine translation [8] to name a few. Though the concept is relatively new, still many enhancements have already been done to further improve the sequence to sequence neural networks. These include: bidirectional models of [3] having encoders that take into account the words in left to right order as well as right to left; the reverse encoder-decoder of [9] that takes the input of the encoder in right to left manner; Character-based models of [10] that feed every encoder not with words in the sentence but with the letter sequence making the sentence. Attention-based models [11], [12] differentiate themselves by bringing into focus the relevant part of the input to the decoder by using the concept of soft attention.

Amongst the diverse set of applications, sequence to sequence networks have gained immense popularity in the domain of machine translation. Many languages are being translated using this method. Google [13] is using the same methodology along with Facebook [14]. [15] have done Chinese-Japanese translation and [16] has shown state-of-the-art results for English to French, English to German amongst many others.

Distributed representation of words (word2vec) in the dataset is another paradigm shift for word representation over using one-hot vectors. Distributed representation was popularized by [5], mapping each word to vector representation that embeds its syntactic and semantic meanings. A variety of hyper parameters needed to be tuned to get the optimal distributed representation amongst which the most important are: the choice between a variant from skip-gram or continuous-bag-of-words; the cut-off for the minimum occurrence of the word; and embedding size. Other algorithms have also been proposed that capture the distributed representation of words out of which Glove [17] and fastText [18] are the most comparable ones. Glove is a count-based model where embeddings are learnt by doing dimensionality reduction. FastText is an extension of word2vec where embeddings for words are represented by the sum of their n-gram embeddings which empowers it to recognize and predict rare words. Favoring the morphologically rich languages, the downside of this property is that FastText does better on learning syntactic embeddings than the semantic embeddings learning [19]. Concept of Word2vec has been extended to sent2vec [20], doc2vec [21], para2vec [21] and topic2vec [22]. These have achieved state-of-the-art results on various natural language processing domains of topic modeling, text summarization and word sense disambiguation. Specifically, a distributed representation of Urdu language has been done by Facebook [18]. However, to the best of our knowledge, these embeddings have not been used for solving any real-life problem. Our distributed word representation is distinct in two ways. First, it is the first Urdu embedding to be used to address a real life problem. Secondly, the dataset is common to the embeddings and the sequence to sequence model; which aligns the embeddings with the dataset resulting in quicker convergence and richer model.

Transliteration is a sub-form of translation such that transliteration changes the letters from one alphabet or language into the corresponding, similar-sounding characters of target alphabet. Specifically, sequence to sequence has been applied to the task of machine transliteration too. [23] transliterates from Sanskrit to English while [12] transliterates from English to Persian using attention based approach. Major challenge of transliteration is addressing the difference in syntax, semantics and morphology between the source language and the target language. There can be issues of misalignment between the source language sentence and the target language sentence and the lengths of both the languages are generally not the same. Needless to mention, these are also the challenges faced for machine translation.

According to [24], Urdu is a low-resource yet morphologically rich and complex language. So far, statistical methods are used to addressing natural language processing in Urdu language [25]. Phrase based statistical machine translation is being done and conventional natural language processing techniques like done POS tagging [24], stemming [26], [24], [27], [28], tokenization, annotation [29], lemmatization [24], sentence boundary detection [24] and named entity recognition [24] are applied in the areas of sentiment analysis, handwriting recognition, opinion mining and plagiarism detection. Work done for the Urdu language has hugely relied on conventional natural language processing techniques [30], [31], [29] and the use of deep learning to address the problem of machine translation in Urdu is still in its inception.

Presence of large scale datasets is an essential requirement for deep learning techniques to work and effectively model the diversity and capture the inherent complexity of language. Availability of parallel corpora opens the avenues for further research for deep learning in machine translation and machine transliteration. For instance, [32] provides a parallel corpora in 11 languages including English, Dutch, Spanish, Italian, and Swedish to name a few. A parallel corpus on a large scale in Urdu language is an avenue unexplored. Absence of Roman-Urdu to Urdu parallel is a bottleneck in exploring further research opportunities in the domain of transliteration as well as translation.

To the best of our knowledge, neural network based machine translation techniques have not been applied for Roman-Urdu to Urdu language transliteration and distributed representation of Roman-Urdu is never explored. Furthermore, no parallel Roman-Urdu to Urdu corpus is available. Thus this work tries to address these shortcomings by providing a large scale parallel corpus for Roman-Urdu to Urdu transliteration with seq2seq model based on distributed representations.

### III. METHODOLOGY

#### A. Model Architecture

Our model is based on Seq2Seq proposed by [9] based on encoders and decoders as illustrated in Figure 1.

The input is a sequence which in our case is a Roman-Urdu sentence and the corresponding output is also a sequence which is in Urdu. Each unit is a LSTM cell which is robust against vanishing gradients and also performs well on longer sequences [33]. As mentioned before, every sentence is in the form of indices where every word in the sentence is represented by a unique index number, given to it in the pre-processing phase.

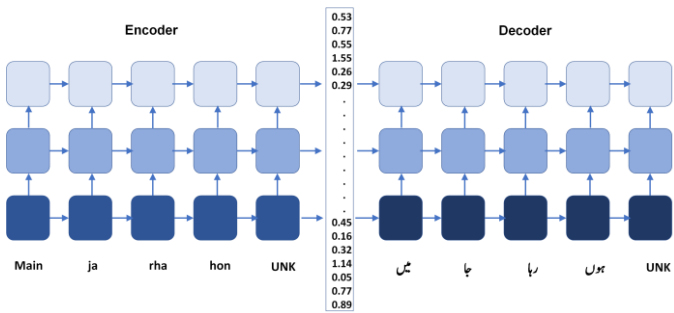

Fig. 1. Seq2Seq network with Roman-Urdu sentence as input to encoder and Urdu sentence as output from decoder.

Likewise, the decoders give the output in the form of indices which point to a unique word in the Urdu vocabulary. It is worth mentioning that the unknown word, 'UNK', is also treated as a word in both Urdu and Roman-Urdu vocabularies as used in [34].

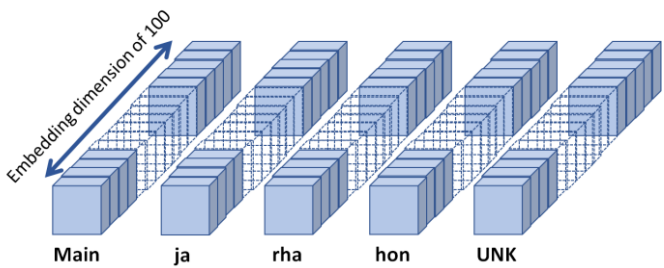

Fig. 2. Distributed representation of 100 dimensional vector built via distributed representation and being fed to a seq2seq network.

We do not give the words or word indices as input but the distributed representation of every word learnt by using the word2vec algorithm proposed by [5]. The way inputs are fed to the network is illustrated in Figure 2. Using distributed representation of words increases the learning capacity of the network and makes faster convergence.

#### B. Data Set

Though Urdu is a morphologically rich language and used by 0.1 billion users, to the best of our knowledge, no work is done in building a publically available Roman-Urdu to Urdu parallel corpus. We collected and built our Roman-Urdu to Urdu dataset as follows:

*1) We* crawled and scraped the web and collected 5.4M Urdu sentences and 0.1M Roman-Urdu sentences. We used only the subset of data collected to transliterate Roman-Urdu sentences to Urdu sentences and vice versa using the web portal [35]. The complete Roman-Urdu to Urdu parallel coprus that we were able to generate had total lines of 0.113 Million.

*2)* We also performed data augmentation to capture the heterogeniety inherent in writing Roman-Urdu for every Urdu word. As an example, the variants of Urdu word رہ in Roman-Urdu can be rha, raha and rahaa. To truly capure the diversity in writing Roman-Urdu, the only solution was to collect data from crowd-sourcing. We could not use publically available data collection sites like Amazon Mechanical Turk since they did not operate in Paksitan nor were the majority of the workers proficient in Urdu language to be able to correctly perform transliteration procedure. However, to fulfill this purpose, we could harness the expertise of 300 people who were proficent in Urdu and Roman-Urdu. It is worth mentioning that it is not feasible to perform transliateration on 5.4M sentences using the above-mentioned particpant strengh. For this, we downloaded 5000 most frequently Urdu words from [36] and 100 variations of each of the 5000 words were collected through crowd-sourcing  Data augmentation was done by using each of the 5000 Urdu words and substituting each word in the data according to its distribution in the data collected through transliteration. Using this technique, a transliterated Urdu to Roman-Urdu parallel corpus with total of 5.3 Million lines was generated.

*3)* We also created a dictionary for one to one mapping between Roman-Urdu and Urdu. We downloaded a word list of 22, 120 Roman-Urdu to Urdu transliterations that we used to create our dictionary. We also expanded our dictionary by using the one to one correspondence between the Roman-Urdu and Urdu in the parallel corpus mentioned above.

#### C. Data Pre-Processing

The following pre-processing steps were applied during transliteration from Roman-Urdu to Urdu.

*1) Removal of noise:* sentences that were abnormally long were removed and the cut off was set to sentence length of 30 words.

*2) Equality of Sentence Lengths:* One-to-one correspondence between the two sentences was needed so that during training, the system learns to accurately map the Roman-Urdu work in place of its Urdu counterpart. Example of the ideal scenario where input and output sequences have the same length is: **Yeh kitab hai** getting transliterated to یہ کتاب ہے.

However, there were many cases when the sequence length in Urdu version is longer than the Roman-Urdu counterpart. For examples, **Yeah Islamabad hai** is transliterated to: یہ اسلام آباد ہے. The Roman-Urdu sequence has length equal to 3 while Urdu sequence has length 4, which is due to the morphological structure of the Urdu script. As one word Islamabad in Roman-Urdu is being mapped to two words in Urdu since there is a space between آباد and اسلام. This leads to the conclusion that one word in Roman-Urdu may get mapped to multiple words in Urdu, which could potentially mislead the model. To overcome this issue, lines of equal length in the both the languages were extracted from the whole corpus, leaving us with the total lines of 0.1M.

*3) Tokenization:* Vocabulary, along with the frequency of each word, was built separately for both the Roman-Urdu and Urdu. Note that the vocabulary size differs despite the total number of words and total number of sentences in both the versions being the same. This is so because for every Urdu word there could be many Roman-Urdu variations. For example, for word یہ, the top five Roman-Urdu variants are yeah, yeh, yeah, ye and yah as mentioned before.

*4) Addition of UNK:* an UNK token is also added to the list of the unique token lists of the vocabularies of the parallel corpora to cater for any unknown word in the parallel corpus. This is also the case for rare words where rare words were replaced by the token UNK to make the system time and memory efficient.

*5) Minimum occurrence of each word:* Using rare words in word2vec leads to a sparse embedding matrix which adds to computational and memory overhead without any gain in performance. One common way to tackle this problem is to put a cap on the vocabulary size. Since high frequency words often provide more information, words with frequency below a certain threshold may assigned as 'UNK' in both the vocabularies. We set the minimum word occurrence as 5 after empirical evaluation. This could be later be replaced by the Urdu word using one to one mapping from Roman-Urdu to Urdu dictionary. Our vocabulary size for Urdu was 19K and for Roman-Urdu it was 20K.

*6)* It is requirement of the Tensorflow's Seq2Seq library that the whole data is input in its indexed form. The same is also true for the output which has to be de-indexed to be human readable. Table 1 explains the process that how Roman-Urdu input is converted to indices during parsing of the input data. In fact, each index is represented by a vector containing its embeddings learnt via word2vec algorithm. Model also outputs its indices where each index represents a word in Urdu language. The word indices for Urdu and Roman-Urdu are different as these indices are assigned on the fly while the data is being read and only at the first occurrence of the word an index is assigned.

TABLE I. INPUTS AND OUTPUTS TO MODEL

| 1 | Our Input | main ja raha hoon |
|---|---|---|
| 2 | Transformed Input | 1003 5 199 7 |
| 3 | Output from Model | 11 987 8 992 |
| 4 | Final Output | میں جا رہا ہوں |

*D. Experimental Settings*

Rigorous experimentation was done on GPU machines to tune the network to identify the most optimal parameters and hyper-parameters.

After thorough experimentation and cross validation procedures, following settings were chosen. i) For learning distributed representations, minimum occurrence of word was chosen as 5 which was needed to filter out the rare words to free the network of extra allocations, and hence led to a sharp decrease in the extra computations. ii) Embedding size of the 100 gave good compromise between performance, computational complexity and memory overhead. This was validated empirically by varying embedding sizes from 5 to 500 with non-uniform intervals. iii) Out of the two variants of word2vec, continuous-bad-of-words version was chosen as it gave better performance on experiments with bigger datasets. iv) Context window size was set to 5, which means for every word to be transliterated 5 words to its left and 5 to its right were considered. These word vectors were generated using the Gensim library [37].

Complete data was randomly shuffled and divided into train (75%) and test set (25%). For the seq2seq architecture, three layer LSTMs were chosen and the maximum sequence length was set to 15. For training, we used Adam optimizer with adaptive learning rate to learn the best parameters of seq2seq with softmax as the loss function.

*E. Results and Analysis*

**Quantitative:**

The evaluation metric BLEU was applied on the train and test sets with varying input sequence lengths. The most optimal sequence length on which train and test both gave the highest score was empirically found to be 10. Our model was able to achieve a BLEU score of 50.68 and 48.6 on train and test data respectively. We can deduce from this small difference in scores that our model is generalizing well and is able to model the contextual and sequential information.

Figure 3 shows the BLEU score against the sentence length. As the sentence length is increased from 2 to 10, the BLEU score shows a gradual increase after which it reduces. From here, we can deduce that the model gives best performance for relatively moderate length sentences. This is true for both train as well as test set. However, it is also evident from the plot that the accuracy gets drastically low after the sentence length gets bigger than 10. It is plausible since inherent architecture of the seq2seq model favors moderate length sentences compared to longer ones.

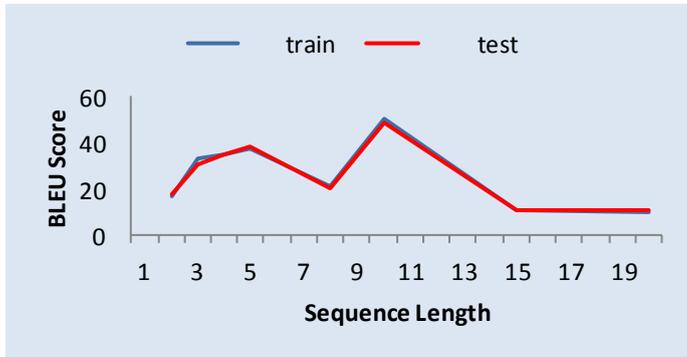

Fig. 3. BLEU score for input sequences of varying lengths.

It is worth noting that during our analysis, we found that no new words are formed and all the transliterated words come from the corpus the model has already seen. This offers advantages for us. Firstly, there is assurance that no unreasonable word is formed unlike the case if the basic unit was taken as the alphabet letters which could have given rise to infinite combination of characters. Secondly, when a new word in Roman-Urdu is encountered due to any of the reasons like a word getting misspelled, shortcuts used in messaging or a slang version of the word, the model looks at the context of the word to be transliterated and comes up with the most probable word. For example, in case of 'mujh per zulmi hova' where the user has misspelled the word zalim as zulmi, the model after having seen many contexts of the zalim as 'mujh', 'per', 'hova' is able to guess the right Urdu word as 'ظلم'. So for the input: '**mujh per zulmi hova**', the output is "مجھ پر ظلم ہوا"

### Qualitative:

Even though the BLEU score appears quite less on numbers, qualitatively we can observe that transliteration done by our model is quite good on both test and train data and it reveals interesting patterns during its transliteration. Three accurate and three erroneous sentences each from train and test set are shown in Table II for analysis and discussion. Words incorrectly transliterated words are highlighted by underlining them. These sentences are randomly chosen from each set and thus give a precise representation of the whole result set.

TABLE II. ACCURATE AND ERRONEOUS RESULTS ON TRAIN AND TEST SETS

| | Train Set | Test Set |
|---|---|---|
| **Accurate Results** | Iqbal musalmanoon ko phir isi akhuwat islami ki taraf lotney ki Talqeen karte hain<br>اقبال مسلمانوں کو پھر اسی اخوت اسلامی کی طرف لوٹنے کی تلقین کرتے ہیں | Inhen bachpan hi se ilm adab aur drama se dilchaspi thi<br>انہیں بچپن ہی سے علم ادب اور ڈرامے سے دلچسپی تھی |
| | Taham dono mumalik ke mabain jung band karwnay ki misbet koshishen nah ho saken<br>تاہم دونوں ممالک کے مابین جنگ بند کروانے کی مثبت کوششیں نہ ہو سکیں | Un ki shairi dil ki kam dimagh ki shairi ziyada hai<br>ان کی شاعری دل کی کم دماغ کی شاعری زیادہ ہے |
| | Is ki wajah shohrat silsila koh namak aur surkh paharhyan bhi hain<br>اس کی وجہ شہرت سلسلہ کوہ نمک اور سرخ پہاڑیاں بھی ہیں | is par ki gayi polish intahi lambay arsay taq apni chamak damak qaim rakhti hai<br>اس پہ کی گئی پالش انتہائی لمبے عرصے تک اپنی چمک دمک قائم رکھتی ہے |
| **Erroneous Results** | Jung mein Meer chaakar Khan ne smh aur bhuto qabail ke mutahidda **mhazon** ko shikast day di<br>جنگ میں میر چاکر خان نے سمہ اور بھٹو قبائل کے متحدہ محاسبہ کو شکست دے دی | Unhon ne **zalim** aur jabir **ohdedaron** ko ohdon se **hataya**<br>انہوں نے ذہانت اور جابر مجرموں کو نظریات والے منگل |
| | Taqreeban saari hi abadi junoob maghribi kinare par **khudaiyon** ke sath abad hai<br>تقریباً ساری ہی آبادی جنوب مغربی کنارے پر کھاکر کے ساتھ آباد ہے | Imla mein koi tabdeeli aisi tajweez nah **ki** jaye<br>املا میں کوئی تبدیلی ایسی تجویز نہ تھی جائے |
| | Tqi sahib phir January ko Bombay ke rastay pani **ke** jahaaz se Karachi aeye<br>تقی صاحب پھر جنوری کو بمبئی کے راستے پانی پر جہاز سے کراچی آئے | Ahal islam ke nazdeek kkhuda se dua **maangna ibadat** mein shaamil hai<br>اہل اسلام کے نزدیک خدا سے دعا قدروں تلاش میں شامل ہے |

Analysis of the transliteration results reveals the following:

1) Observing all the examples collectively, we can deduce that model has aptly learnt that the lengths of both the input and target sentences have to be the same. Even if the words are not predicted correctly, the number of total words in the sentence is the same as that of the input sentence.

2) Model has successfully captured the semantic and syntactic relationships between the source and the target language which is why majority of the words predicted are correct. All sentences make sense and are grammatically and logically correct. It is an encouraging result since Urdu is a complex and morphologically rich language and the dataset used covers all the variety of aspects inherent in the Urdu language. We are able to achieve such accuracy since our parallel corpus is a good mix of simple and difficult words and covers the diversity in grammatical and semantic rules. For example, for the prediction:

انہیں بچپن ہی سے علم ادب اور ڈرامہ سے دلچپی تھی

The words دلچپی , ڈرامہ and بچپن are not very frequently used words. However, the model has accurately predicted them in the context of the overall sentence.

3) Model has learnt the frequently occurring words more accurately than the rare words. This is understandable since the high frequency words give more chance to the model to capture their relationship than the infrequent ones. We can observe that majority of the words not correctly predicted are the infrequently used words like:

نظریات , مجرموں, ذہانت, محاسبہand قدروں .

4) Model does better on shorter sentences than it does on longer ones. Though it is also able to get sentences of length longer than 15 completely correct. The example to support this is the transliteration of "**is par ki gayi polish intahi lambay arsay taq apni chamak damak qaim rakhti hai**" as اس پر کی گئی پالش انتہائی لمبے عرصے تک اپنی چمک دمک قائم رکھتی ہے.

TABLE III. CONTEXTUAL SIMILARITY WHEN PREDICTION IS ERRONEOUS

| Target word | Predicted word | Similarity |
|---|---|---|
| mhazon | محاسبہ | Noun to noun similar pronunciation |
| ke | پر | Preposition to preposition |
| ohdedaron | ذہانت | Noun to noun |
| ibadat | تلاش | Noun to noun |

5) For incorrect predictions, model makes a smart choice by predicting the word which is similar in nature i. e. for a rare word a rare word is predicted, for a preposition, the corresponding preposition is predicted. This is elaborated in Table III. Due to this contextual learning, the mistakes do not get highlighted and the reader fails to notice the mistake when reading is done in a continuous flow.

It is worth mentioning that dictionary-based one-to-one mapping of Roman-Urdu to Urdu dictionary was only successful for seen words not needing any contextual information. However, it failed miserably for: unseen words; words with different connotations; and words requiring contextual information.

Another factor that contributed to the success of these accurate predictions is the use of embeddings for both Urdu and Roman-Urdu vocabularies learnt from the parallel corpora. This also means that we did not create our embeddings from the bigger, non-parallel corpora which despite being richer lacked the preciseness needed by the Seq2Seq model to learn the dependencies. Use of these embeddings led to poor performance in our initial experiments.

IV. CONCLUSIONS

In this paper, we have shown the state-of-the-art results on the problem of Roman-Urdu to Urdu transliteration using sequence to sequence networks with distributed representations. To the best of our knowledge, all the existing transliteration models use traditional rule-based statistical methods and thus are not as generalizable as our proposed deep learning method. The strength of our model is that it is generic, context-aware, and scalable to longer sentences. It gives excellent performance on rare words and out of vocabulary words since it does not rely on one to one mapping of the word rather exploits their contextual information and sequential structure.

V. FOLLOW-UP WORK

We plan on using this model and dataset as the baseline for future work. For instance, performance of our model can be further improved by using the bi-directional encoders with attention mechanism. Similarly, we have not yet explored in detail optimization techniques like bucketing. We leave all this as our future work.